\documentclass[conference]{IEEEtran}
\IEEEoverridecommandlockouts

\usepackage{cite}
\usepackage{amsmath,amssymb,amsfonts}
\usepackage{algorithmic}
\usepackage{graphicx}
\usepackage{textcomp}
\usepackage{xcolor}
\usepackage{multirow}
\usepackage{multicol}
\usepackage{svg}
\usepackage[acronym]{glossaries}
\usepackage{flushend}

\usepackage{subfig}

\newacronym{nms}{NMS}{non maximum suppression}
\newacronym{cnn}{CNN}{convolutional neural network}
\newacronym{hog}{HoG}{histogram of gradient}
\newacronym{PIR}{PIR}{passive infrared}

\def\BibTeX{{\rm B\kern-.05em{\sc i\kern-.025em b}\kern-.08em
    T\kern-.1667em\lower.7ex\hbox{E}\kern-.125em}}

\begin{document}

\title{\emph{Anyone here?}\\ Smart embedded low-resolution omnidirectional video sensor to measure room occupancy\thanks{}}

\author{\IEEEauthorblockN{ \and \and }}

\author{\IEEEauthorblockN{ Timothy Callemein, Kristof Van Beeck and Toon Goedem\'e}
\IEEEauthorblockA{\textit{EAVISE - KU Leuven}\\
Sint-Katelijne-Waver, BELGIUM \\
\{firstname.lastname\}@kuleuven.be}
}

\maketitle

\begin{abstract}
In this paper, we present a room occupancy sensing solution with unique properties: (i) It is based on an omnidirectional vision camera, capturing rich scene info over a wide angle, enabling to count the number of people in a room and even their position. (ii) Although it uses a camera-input, no privacy issues arise because its extremely low image resolution, rendering people unrecognisable. (iii) The neural network inference is running entirely on a low-cost processing platform embedded in the sensor, reducing the privacy risk even further. (iv) Limited manual data annotation is needed, because of the self-training scheme we propose.
Such a smart room occupancy rate sensor can be used in e.g. meeting rooms and flex-desks. Indeed, by encouraging flex-desking, the required office space can be reduced significantly. In some cases, however, a flex-desk that has been reserved remains unoccupied without an update in the reservation system. A similar problem occurs with meeting rooms, which are often under-occupied. By optimising the occupancy rate a huge reduction in costs can be achieved. Therefore, in this paper, we develop such system which determines the number of people present in office flex-desks and meeting rooms. Using an omnidirectional camera mounted in the ceiling, combined with a person detector, the company can intelligently update the reservation system based on the measured occupancy. Next to the optimisation and embedded implementation of such a self-training omnidirectional people detection algorithm, in this work we propose a novel approach that combines spatial and temporal image data, improving performance of our system on extreme low-resolution images.

\end{abstract}


\begin{figure*}[tb]
  \includegraphics[width=\linewidth]{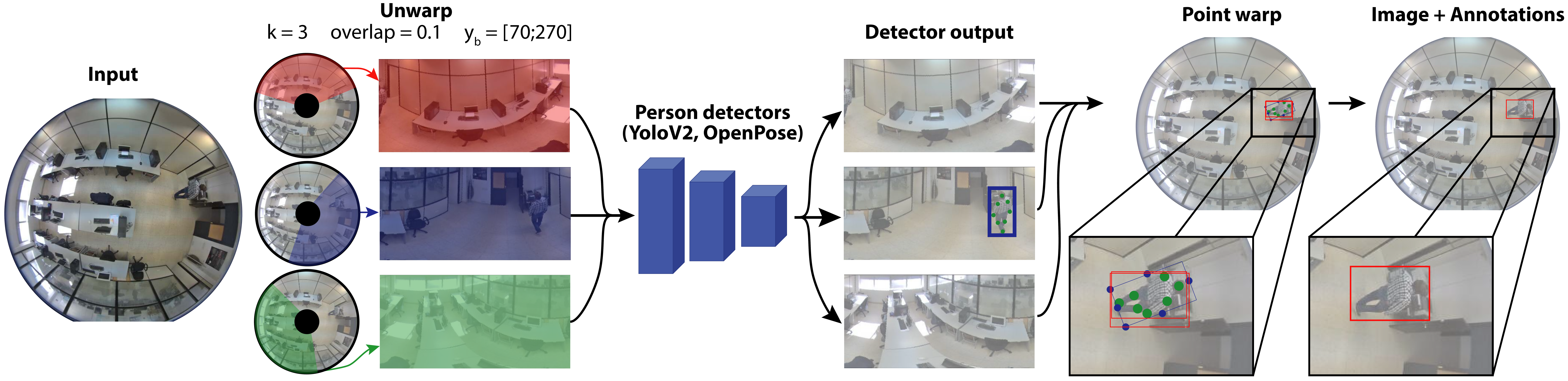}
  \caption{Overview of our proposed self-training approach.}
  \label{fig:approach_generatingboxes}
\end{figure*}

\section{Introduction}
Companies often require larger facilities as their number of employees increases. By encouraging flex-desking, the required office space can be reduced significantly. However, the growing amount of meetings and the rise in popularity of flex-desks in many cases result in building capacity inefficiency problems: reserved meeting rooms and flex desks remain unoccupied due to people working from a different location, cancellations, rescheduling without adjusting the reservation info, and so on.
In other cases, large meeting rooms get reserved, while another smaller meeting room also meets capacity requirements.

To partially resolve this issue, a \gls{PIR} sensor, that measures the change in reflected infrared light, can be used to determine human activity in a room.
However, the sensor proves to be inadequate since enough movement must occur for it to operate and it is unusable to specify the degree of occupancy, i.e. the number of people in the room, or even to determine which desks are taken.
By placing a camera system, more information is gathered that can be analysed with greater accuracy.
Omnidirectional cameras are gaining popularity in security applications because of their wide field-of-view and ease of installation.
Compared to traditional cameras, they capture a 360 degree image without the need of camera re-positioning, which is the case for most traditional motorised (PTZ) cameras.
While they can provide a complete overview at one glance, the images suffer from severe fish-eye lens distortion.
This is not a problem when the camera images are analysed by humans.
However, out-of-the-box state-of-the-art computer vision algorithms which are trained on frontal, upright persons will fail on these images. Hence, we need to retrain such a detector with similar omnidirectional image material.

\begin{figure}[b]
  \includegraphics[width=\linewidth]{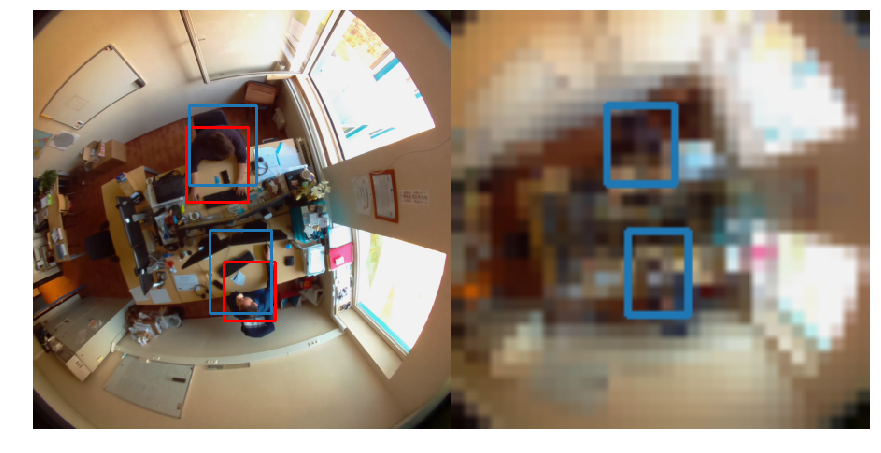}
  \caption{Private dataset example frame, high resolution (left) and low-resolution (right), with automatically generated annotations (red) and detections based on low-resolution image (blue) }
    \label{fig:private_160_32_cp3}
\end{figure}

Unfortunately, the available amount of omnidirectional training data with person annotations is limited.
To overcome this challenge, we propose a self-training approach that uses state-of-the-art person detectors on unwarped omnidirectional data, to automatically generate annotation labels.
These annotation labels are then used as training data to train a new model, capable of determining the room occupancy directly on omnidirectional images. This is illustrated in fig. \ref{fig:private_160_32_cp3}.
In our application, we want to exploit the fact that the ceiling-mounted camera is static and allow the trained models to be environment-specific.

\begin{figure*}[tb]
  \subfloat[High-resolution focused]{
	\begin{minipage}[c][1\width]{
	   0.25\textwidth}
	   \centering
	   \includegraphics[width=\textwidth]{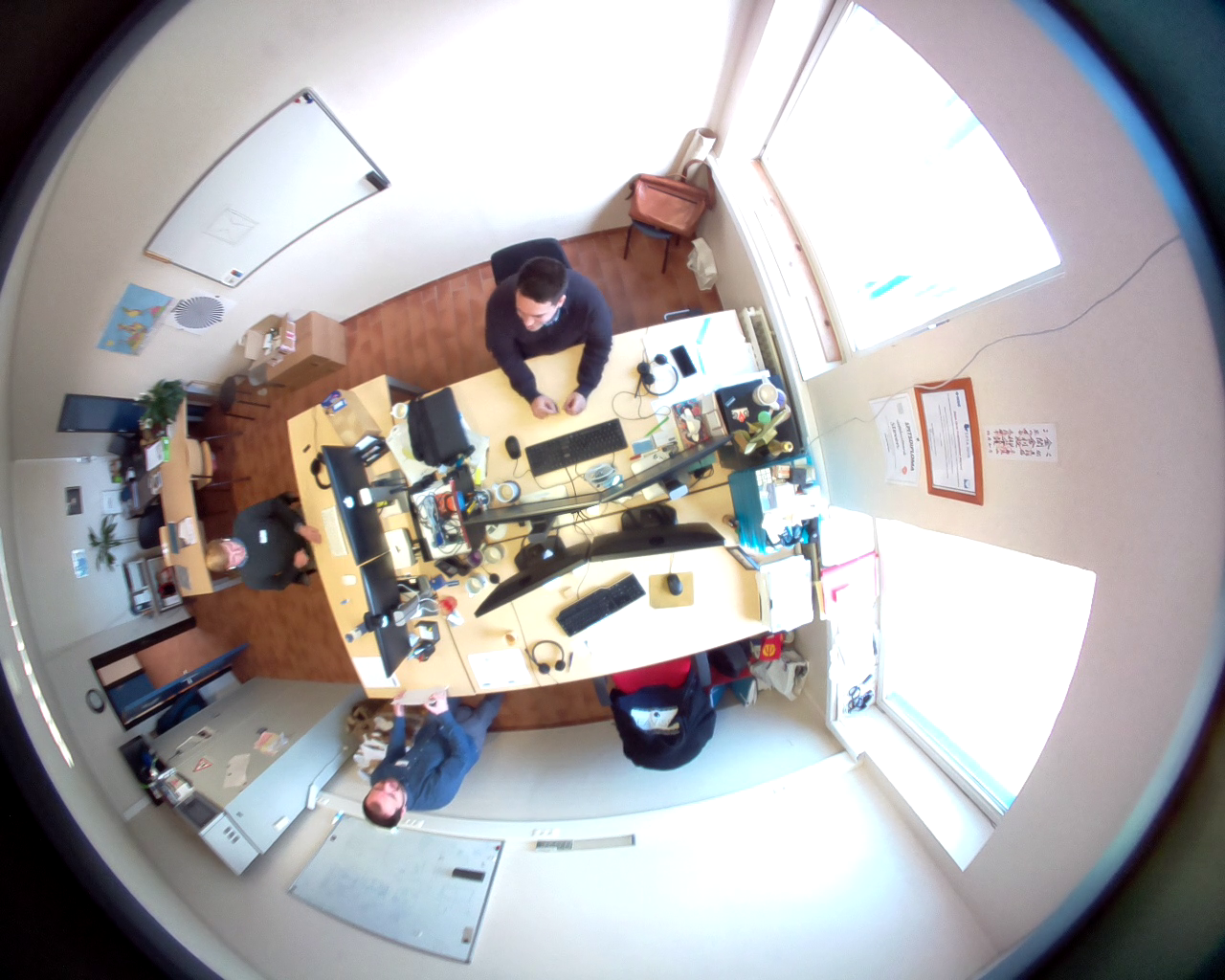}
	\end{minipage}}
  \subfloat[Low-resolution focused]{
	\begin{minipage}[c][1\width]{
	   0.25\textwidth}
	   \centering
	   \includegraphics[width=1\textwidth]{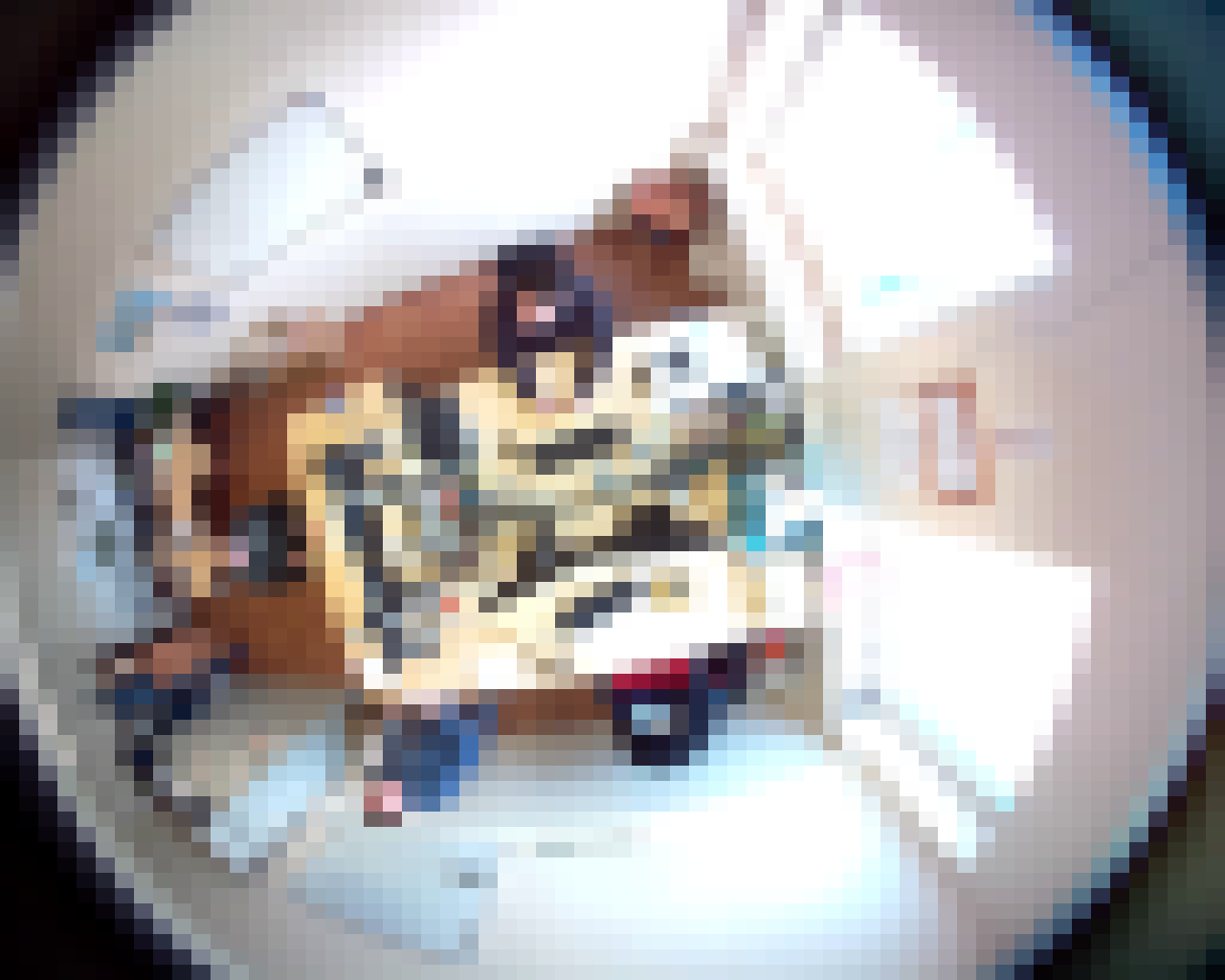}
	\end{minipage}}
  \subfloat[High-resolution out-of-focus]{
	\begin{minipage}[c][1\width]{
	   0.25\textwidth}
	   \centering
	   \includegraphics[width=\textwidth]{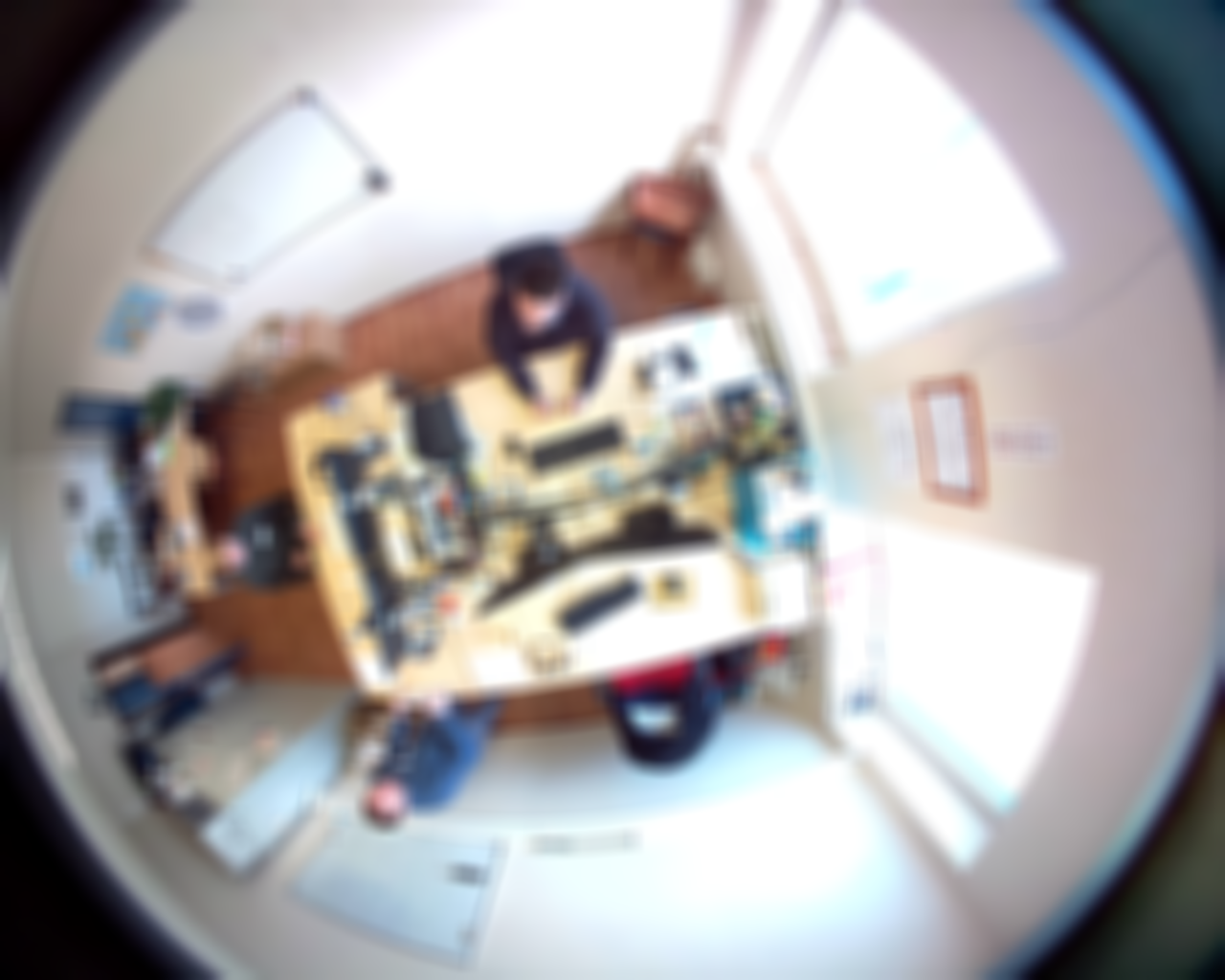}
	\end{minipage}}
  \subfloat[Low-resolution out-of-focus]{
	\begin{minipage}[c][1\width]{
	   0.25\textwidth}
	   \centering
	   \includegraphics[width=\textwidth]{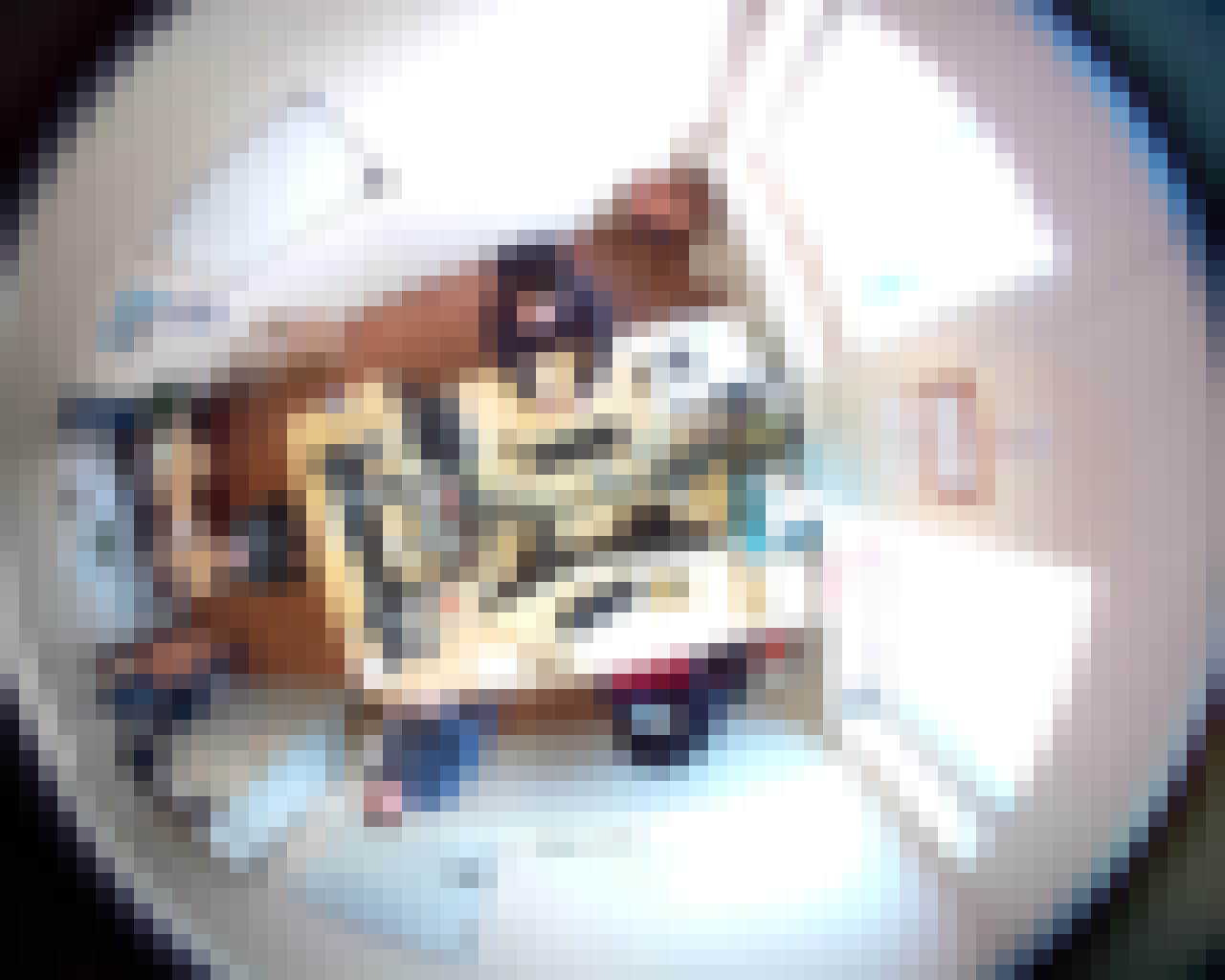}
	\end{minipage}}
\caption{Example camera frames, one high and low resolution frame (64 px) for a frame with a lens in focus, another with a out of focus lens}
\label{fig:in-and-out-of-focus}
\end{figure*}

An eternal issue with camera-based sensors is the concern of people's privacy. 
Indeed, most employees do not feel comfortable when being constantly observed by cameras, and in many cases recording identifiable people in their work environment is not allowed due to legal regulations.
In the system we propose, the privacy is guaranteed because of two reasons. 
Firstly, these privacy issues are avoided if the sensitive information is processed locally (for example on an embedded platform), and only the anonymous meta-data is outputted. 
Our application therefore will be optimized such that it is capable of running on an embedded platform, e.g. a Raspberry Pi.
Secondly, and most importantly, our resulting system works on extreme low image resolution data, in which people are inherently unable to be recognised.
The work by Butler \emph{et al.} \cite{butler2015privacy} supports this, indicating that the sense of privacy is increased when the image contains less details (for example, by lowering the resolution).
After the self-training step, we can even turn the omnidirectional lens out-of-focus, yielding identical downscaled low-resolution input images, but making hacking the sensor purposeless.
Figure \ref{fig:in-and-out-of-focus} shows example frames, the two leftmost frames showing the high and low resolution frame with a lens in focus, and the two rightmost frames with a lens places out of focus.
Apart from regulations, the awareness of being recorded can be considered obtrusive and induce the feeling of being watched and monitored.
Yet by designing the outer shape of the sensor to not resemble a camera, this feeling is greatly reduced.

For the above reasons, we propose to train a state-of-the-art object detector (YOLOv2 \cite{redmon2016yolo9000}) on low-resolution omnidirectional data.
First, we lower the resolution and needed computational power by decreasing the network resolution.
To overcome network architectural limitations, an additional effort was made to reduce the image resolution further.
While lowering the resolution increases privacy, the loss in data will increase the challenge of to accurately detect the room occupancy.
We therefore propose a novel approach that incorporates temporal information to compensate the loss in spatial information.
Towards this goal, we train several object detection models with several different input and image augmentation settings.
In our application, where we aim to count the degree of occupancy, and therefore pay less attention to the location and bounding box output of our models, we use a count-by-detection methodology as end result.

This work goes beyond previous work by Callemein \emph{et al.} \cite{callemein2018low}, with the important novelty that the neural network is implemented on a low-cost embedded device, after several optimisations. 
Moreover, our combination of spatial and temporal image data is clearly boosting the detection performance and further reduce the input resolution as compared to their result.

The remainder of this paper is structured as follows. 
Section \ref{sec:related_work}) discusses the related work followed by section \ref{sec:approach} describing our suggested approach to generating object detection labels and how we can further reduce the image resolution by combining temporal information.
Followed by section \ref{sec:evaluation} where we evaluate our approach on three different datasets.
In section \ref{sec:conclusion} our conclusion with a discussion and possible future work.

\section{Related Work}
\label{sec:related_work}

Object counting has many applications and challenges.
Our case mainly focuses on person counting using omnidirectional camera images. 
Intuitively, we would look at crowd counting architectures capable of estimating the number of people in crowds \cite{arteta2016counting,zhang2016single,sam2017switching}.
Such techniques will train a \gls{cnn} model capable of estimating a crowd density map, based on the head location.
These techniques have no interest in the exact spatial location of the crowd and only focus on estimating the number of people.
However, most crowd counting techniques expect dense crowds on which a relative low error is allowed compared to the high value output.
Our case however, only has a limited sparse number of people (12 max), which is far more sensitive to these errors.

Instead of room occupancy detection using crowd counting techniques, one might also use an object detector and simply count the number of detections. For example, work by Zou \emph{et al.} \cite{zou2017occupancy} uses a two step approach (temporal \gls{hog} head model followed by \gls{cnn} head classifier) to determine the degree of occupancy.
However, their approach does not focus on privacy concerns and uses wall mounted cameras instead which are sensitive to scene occlusions.
Other work by Newsham \emph{et al.} compares a wide variety of sensors mounted on top of a computer screen to determine the occupancy degree (including thermal, \gls{PIR} and radar sensors). 
These sensors avoid the recording of privacy sensitive data.
However, they have several disadvantages: they require installing additional (costly) hardware, enough movement is needed for reliable detection and they are unable to determine the exact degree of occupancy. 
In this work, we aim to remain as unobtrusive as possible, using only a single wide angle ceiling mounted camera.
Therefore, to determine the occupancy degree, we propose to use a vision-based person detector.
Our approach is able to detect persons in extremely low-resolution images, such that the privacy is inherently preserved.

Several deep learning architectures are capable of efficient person detection \cite{ren2015faster,redmon2016yolo9000,liu2016ssd}.
Often the first layers are trained on a large scale dataset (ImageNet \cite{krizhevsky2012imagenet}) and afterwards the full network is fine-tuned for object detection on smaller datasets (VOC \cite{pascal-voc-2012}, COCO \cite{lin2014microsoft}).
Our application will use omnidirectional cameras producing heavily distorted images that are not included in these datasets.
To overcome this challenge, Seidel \emph{et al.} \cite{seidel2018improved} proposes an approach that first transforms the omnidirectional images to perspective images.
On these perspective images they use a person detector, and compare different \gls{nms} approaches to combine the detections that were transformed back on to the original omnidirectional image.
A different approach by Masato \emph{et al.} \cite{tamura2019omnidirectional} tries to train a rotation invariant model by introducing rotation augmentation during training, to partially overcome the rotation distortion of omnidirectional images.
Both works face a similar challenge with only a limited amount of available omnidirectional data.
To overcome this challenge, they either only work on the unwarped image to better fit the model dataset.
Or by rotating the large datasets, to better resemble how people appear on the omnidirectional images.
Our application tries to preserve the privacy by using low-resolution image resolution. 
Unwrapping these low resolution images or using rotated low-resolution images will not use the environment specific data to compensate the loss in data.
By training models on the low-resolution omnidirectional data, we expect the model can better learn what describes the low-resolution representation of people.

Previous work by Callemein \emph{et al.} \cite{callemein2018low} follows this methodology and faces a similar challenge.
They propose an approach to count the number of people present in meeting-rooms and flex desk environments, while working towards privacy preservation.
They also have a limited amount of omnidirectional data suited to their use-case, and therefore recorded their own data in several scenarios.
Since this data was unlabelled, they proposed a teacher-student approach, where the teacher uses Yolov2 \cite{redmon2016yolo9000} and OpenPose \cite{cao2017realtime,cao2018openpose} detector models on unwarped images to first generate labels on their private dataset.

Based on these generated labels they train several Yolov2 models and increase the privacy by reducing the image resolution.
By lowering the resolution they reduce the details that make a person recognisable, increasing the privacy.
However, their teacher pipeline was optimised for their omnidirectional camera and produced large area detections.
Furthermore, they reached a architectural low-resolution limit and only went as low as $96\times96$ pixels.
Our approach proposes flexible detection generation pipeline that produces smaller annotations.
We decreased the resolution further, and propose a novel approach that uses temporal data to retain performance.

\section{Approach}
\label{sec:approach}

We propose a two part approach, capable of counting the amount of people on privacy preserving low resolution omnidirectional data.
The omnidirectional camera will capture a static overview of meeting rooms or flex-desks.
The only data variation is caused by the present people in the room.
We therefore suggest to train a specific model on each scene, instead of generic for multiple scenes.
However, recording new data for each scene requires manual data labelling before the data can be used for training.
To significantly decrease the needed amount of manual labour, we propose an approach capable of autonomously annotating the data, described in section \ref{sec:approach_generatingboundingboxes}.
After autonomously acquiring bounding box labels on the high resolution data, we train several models for extremely low resolutions using these labels.
Lowering the resolution will decrease the image detail and increase the sense of privacy.
However, decreasing image resolution also leads to a significant loss in spatial data.
We therefore propose an approach that is able to retain the model performance using temporal information, described in section \ref{sec:approach_interlacingkernel}.

\subsection{Generating Bounding boxes} \label{sec:approach_generatingboundingboxes}

The high-resolution omnidirectional input image is first unwarped into $k$-images with $overlap$ at either side.
Figure \ref{fig:approach_generatingboxes} shows an example with $k=3$ and an overlap of 10\%.
Additionally we also determine exclusion areas, for example near the heavily distorted centre or upper boundary where people will never be present.
These parameters ($k$, $overlap$, $y_b$) determines the number of unwarped fragments and the width and height of each fragment.
Each fragment will be used as input for both the Yolov2 \cite{redmon2016yolo9000} person detector and OpenPose \cite{cao2018openpose} pose estimator.
Since the unwarping parameters greatly influence the performance of the second step, the optimal settings were determined experimentally, as discussed in section \ref{sec:eval_generatingboundingboxes}.

To warp the detection out on the omnidirectional image, we first transform each bounding box to a poly-point representation.
Instead of only using the bounding box corners, we add 2 evenly separated points on each of the bounding box sides.
The upper left and upper right corners of the bounding box warped on the omnidirectional image will be placed further away from each-other.
When a new bounding box is calculated based on these warped detection points, the top corners will enlarge the detection area greatly.
By removing the top corners of each detection before warping, our warped detection will have a smaller area that better fits around the person.

Both the calculated Yolov2 points and the OpenPose pose estimation output are then warped back on the omnidirectional image.
Around each set of point-detections we calculate a bounding box, and suppress overlapping detections using \gls{nms} with a threshold of 0.4.

\subsection{Interlacing kernel}
\label{sec:approach_interlacingkernel}
Previous work by Callemein \emph{et al.}\cite{callemein2018low} shows it is possible to detect people in similar scenes even after decreasing the image resolution.
However, they are only able to reduce the resolution to 96 pixels (due to architecture limits).
At such resolutions, people arguably remain recognisable.
Our approach allows for extreme low resolutions, not limited by network constraints.

We use the Yolov2 object detection architecture to train several models, using the autonomously generated bounding boxes, discussed in section \ref{sec:approach_generatingboundingboxes}, with a network resolution of 160px and 96px.
Yolov2 uses network resolution resize augmentation to allow the model to learn different scales.
For this purpose, they resize their network between a range of [320; 608] with a step of 32.
We follow a similar approach and allow the network to randomly resize the network resolution within a range of $[net_{res} - 32*2; net_{res} + 32*2]$.
Architectural limitations only allow the lowest network resolution of 96px.
In the case of $net_{res}=96$, the network will only have random upscales and saturate at 96px.
While $net_{res}=160$, is still able to use the full random resize scope.

Apart from decreasing the network resolution, we propose several down-up-scale resolutions (64px, 48px and 32px).
The images are first down-scaled to these extreme low resolutions and up-scaled with linear interpolation to the network resolution.
For each network resolution and down-scale resolution a model was trained to assess its performance.

\begin{figure}[]
  \includegraphics[width=0.95\linewidth]{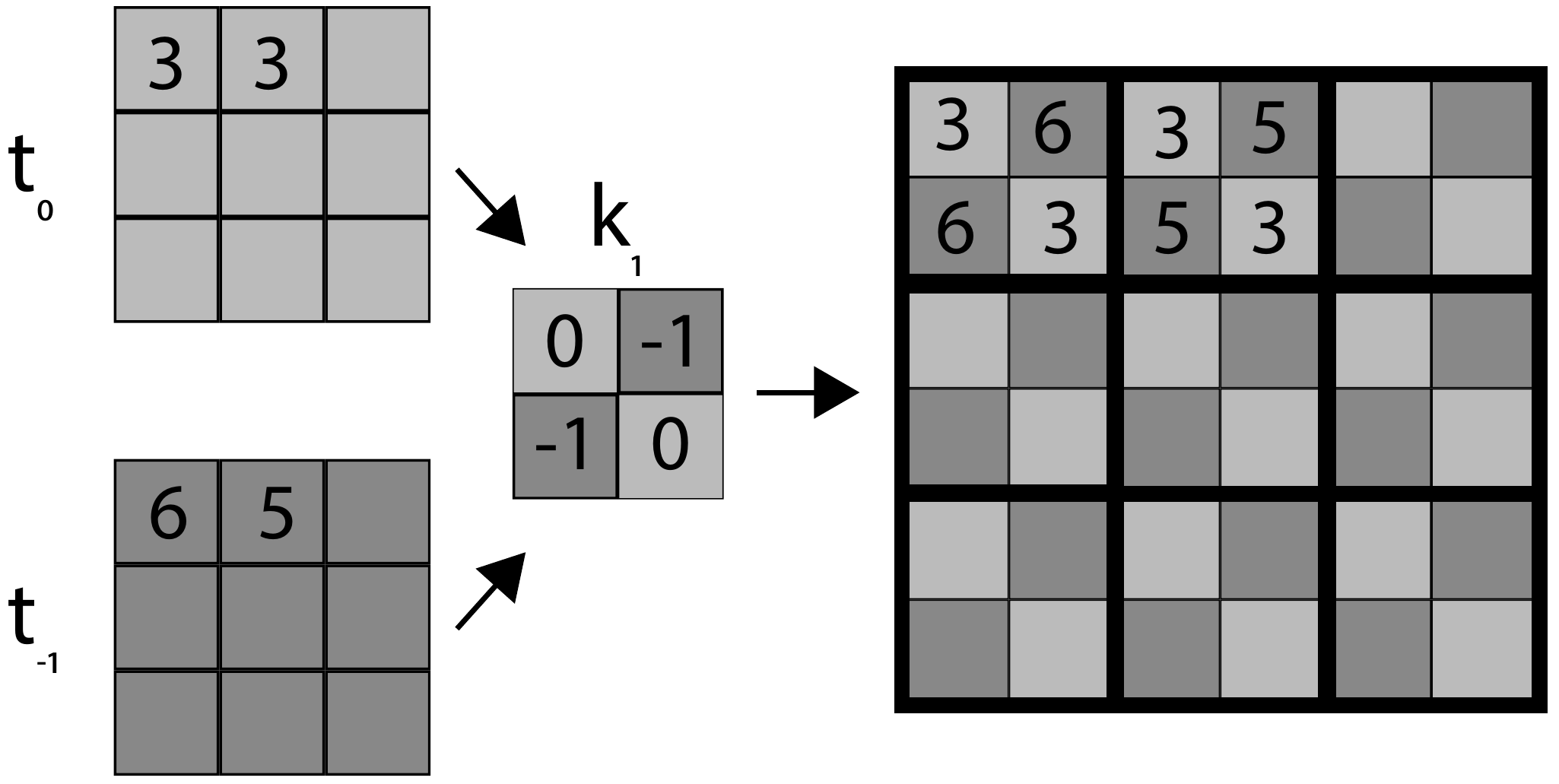}
  \caption{overview showing an example temporal upscale combining multiple frames, resembling interlacing}
  \label{fig:interlacing-example}
\end{figure}

While reducing the resolution further increases the privacy, it will also lead to general detail and data loss, increasing the challenge to detect people.
To cope with this, we expect the region around people to contain temporal information, on the assumption that people constantly move over time, while e.g. furniture remains static.
Based on this assumption, we propose to use this temporal information by combining multiple low-resolution images with \emph{interlacing kernels}.
Interlacing, used in video compression and computer graphics, is a technique where only parts of each frame are stored (reducing the required storage) and shown when watching the video.
These interlaced frames combine multiple low-resolution frames together into a single higher resolution image.
In our case we use a similar concept as interlacing, namely combining temporal-spatial information to increase the image resolution.
This way we can improve performance compared to a linear interpolation up-scale.
We therefore propose to use kernels (further referenced as \emph{interlacing} kernels) to combine multiple frames together.
Figure \ref{fig:interlacing-example} illustrates our proposed approach, combining two $3 \times 3$ matrices into a $6 \times 6$ matrix using a $2 \times 2$ interlacing kernel.
This interlacing kernel contains index values and determines the source matrix to gather pixel data from.
When no movement has taken place, these pixels will show similar behaviour as up-scale interpolation.
In case of movement, we assume that the pixels will contain this movement and show more edges opposed to interpolation output.

\section{Evaluation}
\label{sec:evaluation}

\begin{table*}[]
\centering
\begin{tabular}{|c|c|c|c|c|c|c|}
\hline
\multicolumn{2}{|c|}{\textbf{Dataset}}      & \textbf{Frames} & \textbf{Sequence}     & \textbf{Dataset Annotation type}          & \multicolumn{1}{c|}{\textbf{People}} & \textbf{Level of movement} \\ \hline
\multirow{4}{*}{Mirror Challenge} & train a & 1084            & 2, 3                  & \multirow{2}{*}{public bounding boxes}    & \multirow{4}{*}{3}                   & \multirow{4}{*}{moderate}  \\
                                  & train b & 3608            & 7,8,10,12,13,14,17,18 &                                           &                                      &                            \\ \cline{5-5}
                                  & test a  & 1123            & 1,4,5                 & \multirow{2}{*}{manually annotated bounding boxes} &                                      &                            \\
                                  & test b  & 2246            & 9,11,15,16            &                                           &                                      &                            \\ \hline
\multirow{3}{*}{PIROPO}           & train a & 10969           & omni\_1a              & \multirow{3}{*}{head points}              & \multirow{3}{*}{3}                   & \multirow{3}{*}{high}      \\
                                  & train b & 4585            & omni\_1b              &                                           &                                      &                            \\
                                  & test b  & 1181            & test1, test2, test3   &                                           &                                      &                            \\ \hline
\multirow{2}{*}{internal office}  & train   & 7686            &                       & None                                      & \multirow{2}{*}{4}                   & \multirow{2}{*}{limited}   \\ \cline{5-5}
                                  & test    & 9953            &                       & manually annotated bounding boxes                  &                                      &                            \\ \hline
\end{tabular}
\vspace{0.2cm}
\caption{Used datasets during our experiments, showing the aggregated sequences, the number of frames, people and movement level.}
\label{tbl:dataset-summary}
\end{table*}

To evaluate both our autonomous label generation, described in section \ref{sec:approach_generatingboundingboxes}, and our proposed approach to combine spatial-temporal data, described in section \ref{sec:approach_interlacingkernel}, we use two publicly available datasets, PIROPO\footnote{https://www.gti.ssr.upm.es/research/gti-data/databases} and MirrorChallenge\footnote{https://www.hcd.icat.vt.edu/mirrorworlds/challenge/index.html}.
Both the PIROPO and MirrorChallenge dataset contain multiple camera setups and positions, we only use the open space and flex desk sequences since they better fit our case.
In order to further test our system and simulate real office space situations, we recorded a private office dataset with little movements since the people are at their desks.
Table \ref{tbl:dataset-summary} shows the summary of used datasets during our experiments.

Section \ref{sec:eval_generatingboundingboxes} evaluates the proposed approach to generate bounding box labels on all three training datasets.
Based on the best settings, we will then use these automatically generated labels to train several models, evaluated on the test datasets in section \ref{sec:eval_interlacingkernel}.

\subsection{Automatic labelling}
\label{sec:eval_generatingboundingboxes}
We can only evaluate our automatic bounding box generation technique on both the MirrorChallenge and PIROPO training datasets, since our private office dataset has no training labels.
The PIROPO dataset, however, has no bounding box annotations, only head point annotations.
We therefore use point-wise evaluation and checking whether the detection box contains the head annotation point.
When this is the case, we annotate it as a true positive, when the point is outside any of the detection boxes it is counted as a false negative.
The remainder of the detections that was not matched with a head annotation are counted as false positives.

\begin{figure}[]
  \includegraphics[width=\linewidth]{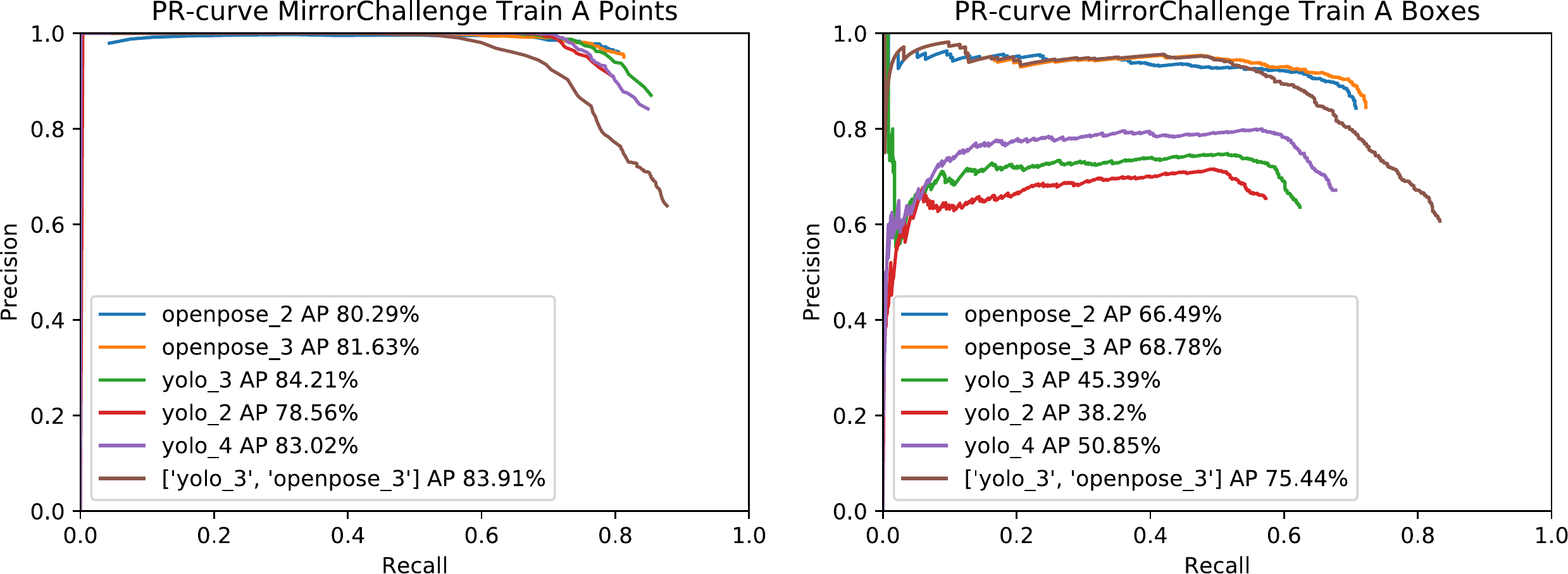}
  \caption{Mirror Train A generated annotations PR-curves evaluating the boxes with an IoU=0.4 and point within bounding box.}
  \label{fig:mirror_a_pr}
\end{figure}

\begin{figure}[]
  \vspace{-0.3cm}
  \includegraphics[width=\linewidth]{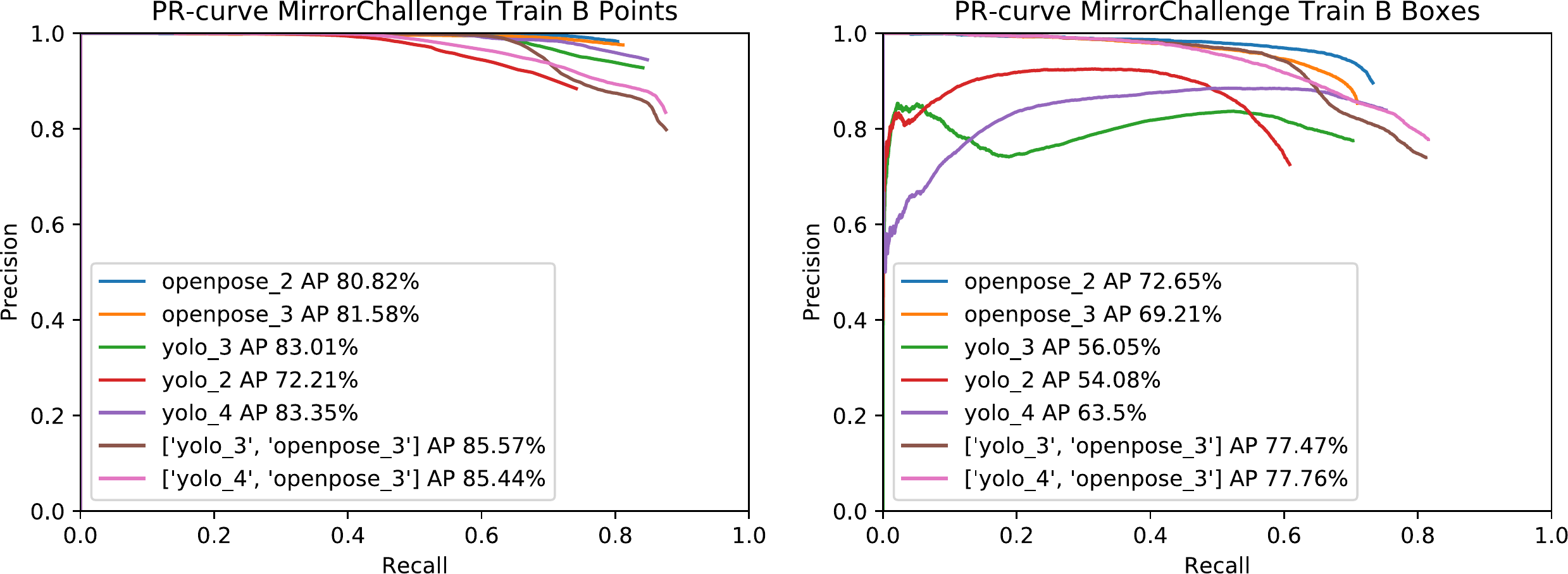}
  \caption{Mirror Train B generated annotations PR-curves evaluating the boxes with an IoU=0.4 and point within bounding box.}
    \label{fig:mirror_b_pr}
\end{figure}

\begin{figure}[]
  \includegraphics[width=1\linewidth]{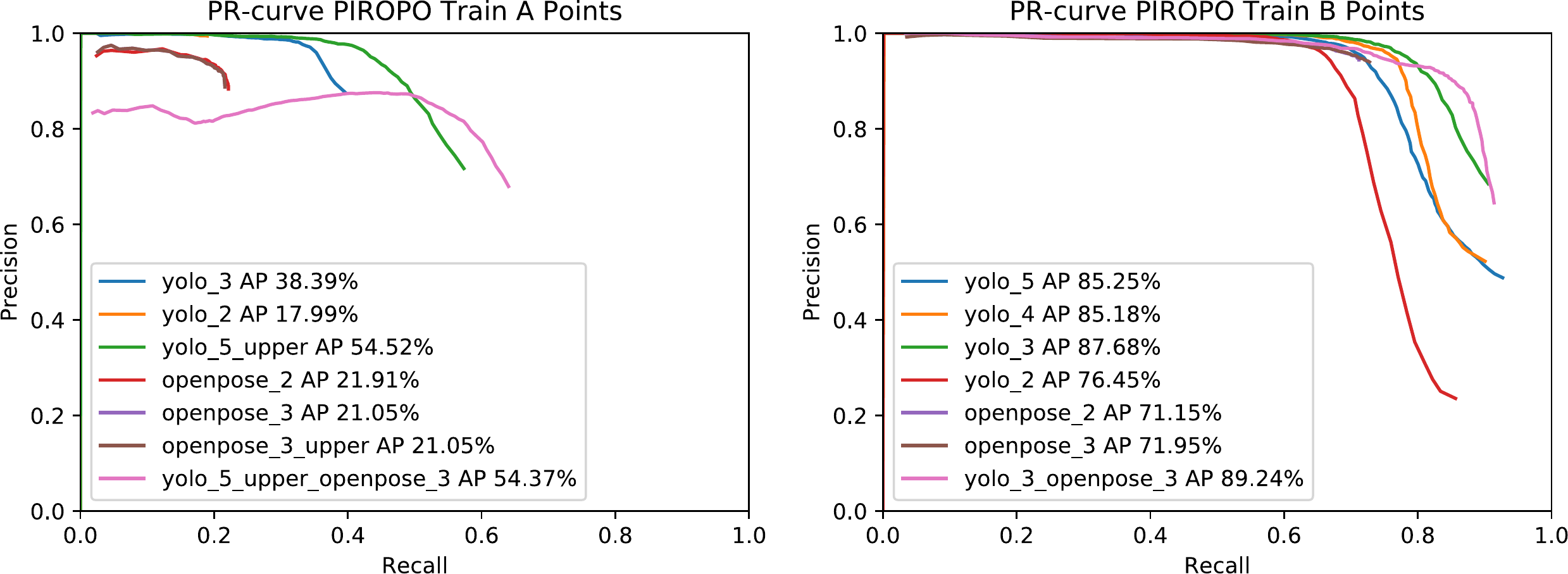}
  \caption{PIROPO Train A and B generated annotations PR-curves evaluating point within bounding box.}
    \label{fig:piropo_pr}
\end{figure}

Figures \ref{fig:mirror_a_pr} and \ref{fig:mirror_b_pr} illustrate two precision-recall curves for the training set A and B of the MirrorChallenge dataset.
The leftmost pr-curve shows the results when using point-wise evaluation, the rightmost will compare the bounding boxes with an IoU of 0.4.
As mentioned in section \ref{sec:approach_generatingboundingboxes} different settings can be used to generate the bounding boxes.
In our case we evaluated different amount of k-frames, with $k = [2;3;4]$ for YoloV2 and $k=[2;3]$ for OpenPose.
Since the PIROPO training datasets only have point annotations, figure \ref{fig:piropo_pr} only illustrates the pr-curves using point evaluation, with $k = [2;3;4;5]$ for YoloV2 and $k=[2;3]$ for OpenPose.
On the PIROPO train set A, we noticed that most of the head annotations were near the circular boundary of the omnidirectional image.
We therefore set $y_b$ to only use the upper-part when unwarping the omnidirectional image, producing fragments with a large width and a small height.
By increasing $k$, we improve the width/height ratio to better fit our detection architectures, resulting in higher accuracy.
The best performing OpenPose and YoloV2 detections are then combined using \gls{nms}.
To further increase the accuracy, we compare the number of detections with the mean number of detections of the past.
If the current frame has a number of detections not equal to the mean number, we allow the system to drop the current frame.
We then determine the optimal threshold with the F1 score and use this threshold to generate annotations that were used for training the models described in section \ref{sec:approach_interlacingkernel}.

To test whether the automatically generated annotations are adequate enough, we trained three models with different network resolutions [448; 160; 96] for each training dataset that were evaluated on the test sets.
In table \ref{tbl:three-network-resolutions} you can find the average precisions, showing good results on the original 448 resolution, and a slight decrease after decreasing the network resolution.
This shows that we are capable of training models with our generated annotations with acceptable performances.

\begin{table}[]
\centering
\begin{tabular}{|c|c|c|c|}
\hline
\multirow{2}{*}{\textbf{Dataset}} & \multicolumn{3}{c|}{\textbf{Net resolution}} \\ \cline{2-4} 
                                  & \textbf{448}  & \textbf{160}  & \textbf{96}  \\ \hline
PIROPO Test B                     & 0.966         & 0.817         & 0.428        \\
MirrorChallenge Test A            & 0.878         & 0.670         & 0.676        \\
MirrorChallenge Test B            & 0.942         & 0.911         & 0.864        \\
Private Office                    & 0.967         & 0.831         & 0.791        \\ \hline
\end{tabular}\vspace{0.2cm}
\caption{Average precision results for each network input resolution, trained on the training sets with automatically generated labels and evaluated on the test sets.}
\label{tbl:three-network-resolutions}
\end{table}

\begin{table}[]
\centering
\begin{tabular}{|c|c|c|c|c|c|c|}
\hline
\multirow{2}{*}{\textbf{Dataset}} & \multirow{2}{*}{\textbf{Net Res}} & \multirow{2}{*}{\textbf{Scale res}} & \multirow{2}{*}{\textbf{Linear}} & \multicolumn{3}{c|}{\textbf{Interlacing Kernel}}  \\ \cline{5-7} 
                                  &                                   &                                         &                                   & \textbf{k1}    & \textbf{k2}    & \textbf{k3}    \\ \hline
                                  & \multirow{3}{*}{160}               & 64                                      & 0.746                             & 0.824          & \textbf{0.860} & 0.655          \\
                                  &                                   & 48                                      & 0.745                             & 0.757          & \textbf{0.833} & 0.750          \\
PIROPO                            &                                   & 32                                      & 0.759                             & 0.774          & \textbf{0.853} & 0.748          \\ \cline{2-7} 
TEST B                            & \multirow{2}{*}{96}               & 48                                      & 0.351                             & 0.428          & \textbf{0.526} & 0.323          \\
                                  &                                   & 32                                      & 0.271                             & \textbf{0.306} & 0.232          & 0.135          \\ \hline
                                  & \multirow{3}{*}{160}               & 64                                      & 0.575                             & \textbf{0.758} & 0.648          & 0.717          \\
                                  &                                   & 48                                      & 0.636                             & \textbf{0.806} & 0.686          & 0.698          \\
MIRROR                            &                                   & 32                                      & 0.585                             & \textbf{0.673} & 0.552          & 0.464          \\ \cline{2-7} 
TEST A                            & \multirow{2}{*}{96}               & 48                                      & 0.435                             & 0.191          & \textbf{0.610} & 0.126          \\
                                  &                                   & 32                                      & 0.557                             & 0.169          & \textbf{0.707} & 0.257          \\ \hline
\multirow{2}{*}{}                 & \multirow{3}{*}{160}               & 64                                      & \textbf{0.939}                    & 0.897          & 0.885          & 0.866          \\
                                  &                                   & 48                                      & \textbf{0.926}                    & 0.895          & 0.906          & 0.864          \\
MIRROR                            &                                   & 32                                      & 0.916                             & 0.882          & \textbf{0.917} & 0.877          \\ \cline{2-7} 
TEST B                            & \multirow{2}{*}{96}               & 48                                      & \textbf{0.877}                    & 0.859          & 0.869          & 0.814          \\
                                  &                                   & 32                                      & \textbf{0.872}                    & 0.850          & 0.838          & 0.852          \\ \hline
                                  & \multirow{3}{*}{160}               & 64                                      & \textbf{0.866}                    & 0.651          & 0.793          & 0.797          \\ 
                                  &                                   & 48                                      & 0.751                             & 0.825          & \textbf{0.879} & 0.641          \\ 
PRIVATE                           &                                   & 32                                      & \textbf{0.710}                    & 0.520          & 0.616          & 0.439          \\ \cline{2-7} 
TEST                              & \multirow{2}{*}{96}               & 48                                      & 0.641                             & 0.583          & 0.639          & \textbf{0.669} \\ 
                                  &                                   & 32                                      & \textbf{0.66}                     & 0.583          & 0.652          & 0.392          \\ \hline
\end{tabular}
\vspace{0.2cm}
\caption{PIROPO, MIRROR A, MIRROR B and Office average precisions on different network and down-scale resolution comparing linear upscale vs. upscale with interlacing kernels.}
\label{tbl:interlacing-results}
\end{table}

\begin{table}[]
\centering
\begin{tabular}{|c|c|c|c|c|c|}
\hline
\multirow{2}{*}{\textbf{Net Res}} & \multirow{2}{*}{\textbf{Scale Res}} & \multirow{2}{*}{\textbf{Linear}} & \multicolumn{3}{c|}{\textbf{Interlacing Kernel with delta t}} \\ \cline{4-6} 
                                  &                                         &                                   & \textbf{1}   & \textbf{2}        & \textbf{3}      \\ \hline
\multirow{3}{*}{160}              & 64                                      & 0.866                             & 0.793        & \textbf{0.878}    & 0.844           \\
                                  & 48                                      & 0.751                             & 0.879        & 0.794             & \textbf{0.892}  \\
                                  & 32                                      & 0.710                             & 0.616        & \textbf{0.7836}   & 0.741           \\ \hline
\end{tabular}
\vspace{0.2cm}
\caption{Results on the private dataset, using kernel $k_2^t$ for $t=[1;3]$.}
\label{tbl:interlacing-timedelta}
\end{table}

\subsection{Interlacing kernel}
\label{sec:eval_interlacingkernel}

The main focus of this paper is counting the people present in omnidirectional images, while lowering the resolution to increase the privacy preservation.
Note that our purpose is to make people detectable, but not identifiable.
Thus the absolute size of the pixel has less importance than the relative pixel size to the size of the face.
However, in our case, we use fixed ceiling mounted top down looking cameras with a wide angle lens adding lens distortion.
This implies that the size of the persons relative to the overall camera image is already small. 
During evaluation we will use point evaluation better evaluating the counting output, while this still takes in account the rough location of the detection.

\begin{figure}[tb]
  \centering
  \includegraphics[width=0.75\linewidth]{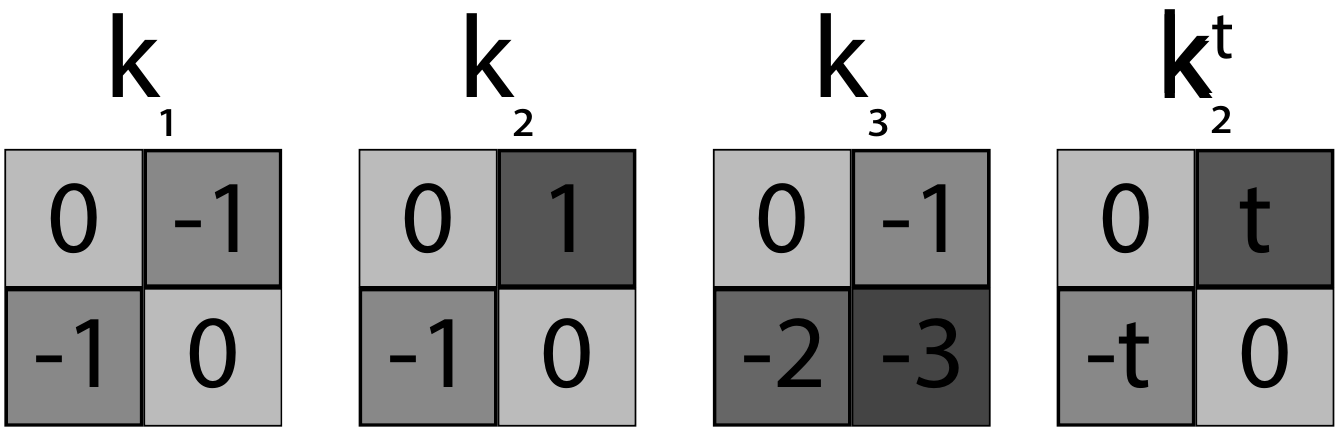}
  \caption{Proposed interlacing kernels used to up-scale the low-resolution images.}
  \label{fig:interlacing_kernels}
    \vspace{1em}
\end{figure}

Table \ref{tbl:interlacing-results} show the average precisions for all trained models on each dataset, with a network resolution of 160 and 96, on images that were down-scaled to a smaller resolution.
We compare linear interpolation with three temporal interlacing up-scaling kernels $k_1; k_2; k_3$.
The three leftmost kernels illustrated in figure \ref{fig:interlacing_kernels} were used, where the value represents the frame index time difference to be used to upscale the images, the datasets are all recorded at 15 fps.
A time-delta of $1$ will result in a 66ms shift in time.
We observe that the effect of taking time into account by using these interlacing kernels depends on the amount of movement of the people in the images. People walking around indeed will generate much more interlacing artefacts than people sitting immobile at their desks.
The results on the PIROPO test sets, with high level of movement, Table \ref{tbl:interlacing-results} shows that in all cases this model outperforms up-scaling the low-resolution images using linear interpolation.
On the MirrorChallenge test sets, with a moderate level of moment, the interlaced upscale does not always outperform the linear interpolation models, but shows similar results.
On our private office dataset, with a very limited level of movement, the same conclusion as on the MirrorChallenge counts, and both perform well.
Yet, the interlaced up-scale in some cases shows a little increase or decrease in performance.

On our private office dataset a smaller impact of time with kernel $k_2$ is visible and to be expected, since little variation occurs in a 132ms time window.
The kernel base-structure as $k_2$ was used, adjusted to hold a variable time-delta $t$, kernel $k_2^t$, illustrated in the rightmost kernel in figure \ref{fig:interlacing_kernels}.
The results for models trained with $k_2^t$ with $t = [1,2,3]$ are show in table \ref{tbl:interlacing-timedelta}, showing that increasing the time-delta increases the performance and outperforms the linear interpolation which was used as the baseline.

\begin{figure}[tb!]
  \subfloat[MIRROR TEST A]{
	\begin{minipage}[c][0.43\width]{
	   \linewidth}
	   \centering
	   \includegraphics[width=0.9\textwidth]{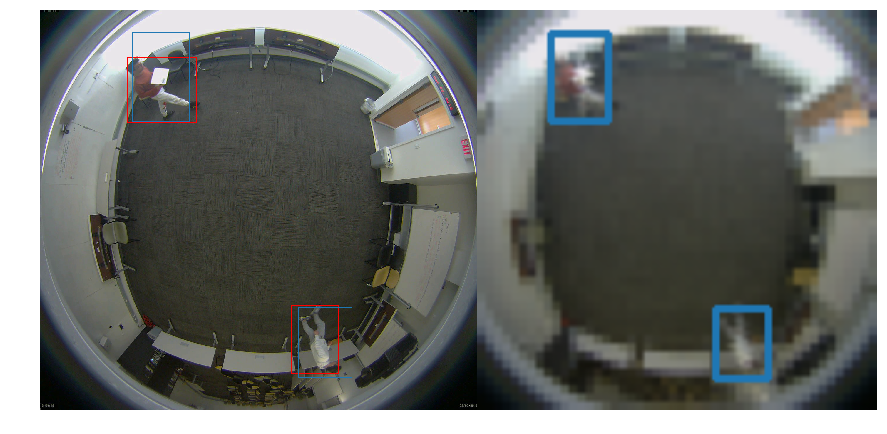}
	\end{minipage}}
	\\
  \subfloat[MIRROR TEST B]{
	\begin{minipage}[c][0.43\width]{
	   \linewidth}
	   \centering
	   \includegraphics[width=0.9\textwidth]{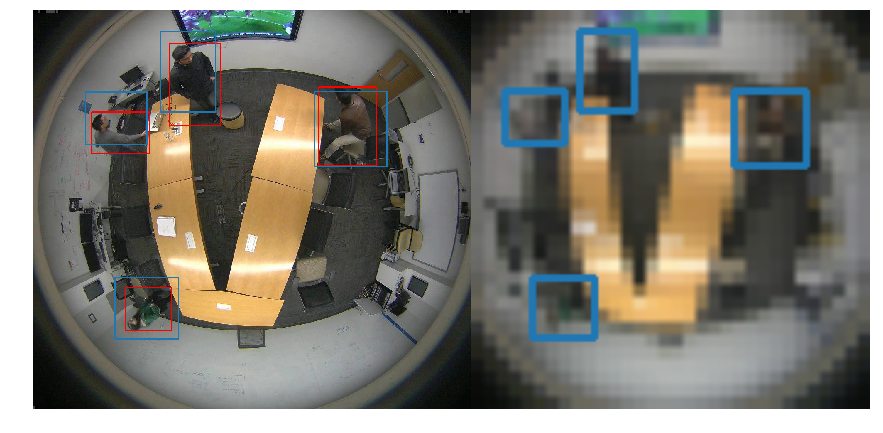}
	\end{minipage}}
	\\
  \subfloat[PIROPO TEST B]{
	\begin{minipage}[c][0.43\width]{
	   \linewidth}
	   \centering
	   \includegraphics[width=0.9\textwidth]{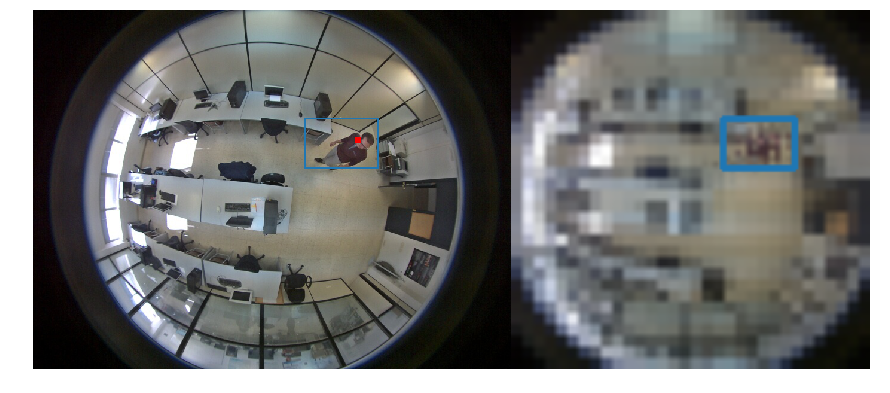}
	\end{minipage}}
	\\
  \subfloat[PRIVATE TEST]{
	\begin{minipage}[c][0.43\width]{
	   \linewidth}
	   \centering
	   \includegraphics[width=0.9\textwidth]{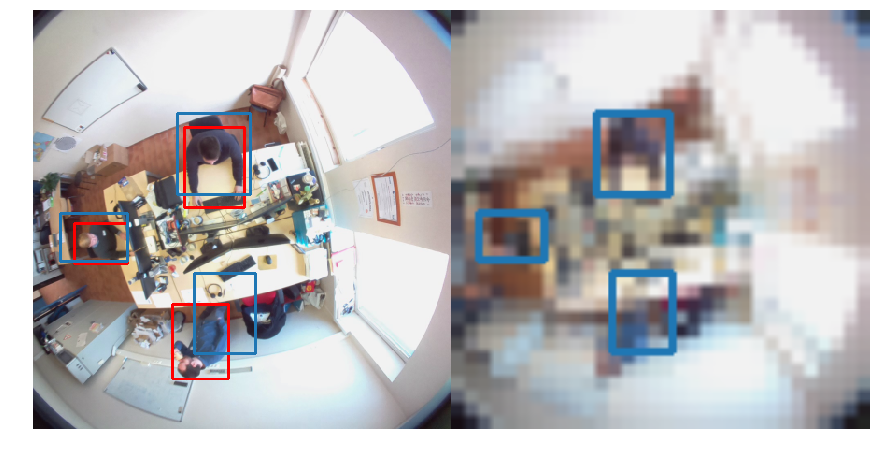}
	\end{minipage}}
\caption{Dataset examples, showing the annotations (red) and detections (blue) on both the high and low resolution frames (Net Res: 160 Scale Res: 32 with interlacing kernel $k_2$).}
\label{fig:four_dataset_output}
\end{figure}

On our private dataset, table \ref{tbl:three-network-resolutions} shows the average precisions using our baseline, which is the approach of Callemein \emph{et al.} \cite{callemein2018low}.
Table \ref{tbl:interlacing-timedelta} shows the results with our proposed approach on the same dataset.
We indeed see that our newly proposed approach enables to reduce the image resolution drastically further (three times lower as in \cite{callemein2018low}), while keeping the average detection precision similar.
Figure \ref{fig:four_dataset_output} shows example images from the each test set, showing the ground truth (red) and the low-resolution based detections (blue) on both original high-resolution image (left) and the low-resolution image (32px with interlacing kernel $k_2$).

\subsection{Embedded implementation}
We implemented the resulting networks on Raspberry Pi 2, 3, and 3B devices, sporting a RaspiCam camera with a 1.1 mm omnidirectional lens. 
Table \ref{tbl:processing-speed-embedded} shows the measured frame rates achieved by our models on these embedded platforms, after automatic self-training.

\begin{table}[tb]
\vspace{1em}
\begin {center}
\begin{tabular}{|c|c|c|}
\hline
\textbf{Device}                   & \multicolumn{1}{c|}{\textbf{Resolution}} & \textbf{Seconds per Frame} \\ \hline
\multirow{3}{*}{Raspberry Pi 2}   & 448                                      & 18.60                      \\
                                  & 160                                      & 3.60                       \\
                                  & 96                                       & 2.17                       \\ \hline
\multirow{3}{*}{Raspberry Pi 3B}  & 448                                      & 16.60                      \\
                                  & 160                                      & 2.96                       \\
                                  & 96                                       & 1.83                       \\ \hline
\multirow{3}{*}{Raspberry Pi 3B+} & 448                                      & 11.72                      \\
                                  & 160                                      & 2.07                       \\
                                  & 96                                       & 1.30                       \\ \hline
\end{tabular}
\end{center}
\caption{Processing speed of our models on embedded platforms.}
\label{tbl:processing-speed-embedded}
\end{table}

A measurement update rate of 1.3s to 4s does maybe not seem enormous, but undeniably it is very suited for the application at hand.
The number of people in a room must certainly not measured more frequently for such a room reservation system.

\section{Conclusion}
\label{sec:conclusion}
In this paper, we presented a omnidirectional camera-based sensor counting the number people in flex-desk and meeting room environments.
To overcome the scarce amount of labelled omnidirectional data an autonomous label generation system based on state-of-the-art person detectors was proposed, allowing for scene-specific data recording, label generation and training of several models.
By decreasing the image resolution we achieve truly privacy-preservation, reducing the input image resolution to the utmost.
To retain high detection accuracy, we proposed to incorporate temporal data to compensate for the loss in spatial data.
Results showed that our approach is capable of using scene knowledge to generate labels that can be used during training.
Evaluating the models, trained on these generated labels, showed that our generated labels were adequate enough to train models capable of counting people with high accuracy.
Furthermore, by using interlacing kernels that take in account temporal information, we see a clear improvement over normal interpolation up-scaling techniques.

\bibliography{biblio.bib} 

\begin{thebibliography}{10}

\bibitem{butler2015privacy}
D.~J. Butler, J.~Huang, F.~Roesner, and M.~Cakmak, ``The privacy-utility
  tradeoff for remotely teleoperated robots,'' in {\em ACM/IEEE International
  Conference on Human-Robot Interaction}, pp.~27--34, ACM, 2015.

\bibitem{redmon2016yolo9000}
J.~Redmon and A.~Farhadi, ``Yolo9000: Better, faster, stronger,'' {\em arXiv
  preprint arXiv:1612.08242}, 2016.

\bibitem{callemein2018low}
T.~Callemein, K.~Van~Beeck, and T.~Goedem{\'e}, ``How low can you go?
  privacy-preserving people detection with an omni-directional camera.,'' in
  {\em International Joint Conference on Computer Vision, Imaging and Computer
  Graphics Theory and Applications.}, International Joint Conference on
  Computer Vision, Imaging and Computer~…, 2018.

\bibitem{arteta2016counting}
C.~Arteta, V.~Lempitsky, and A.~Zisserman, ``Counting in the wild,'' in {\em
  European conference on computer vision}, pp.~483--498, Springer, 2016.

\bibitem{zhang2016single}
Y.~Zhang, D.~Zhou, S.~Chen, S.~Gao, and Y.~Ma, ``Single-image crowd counting
  via multi-column convolutional neural network,'' in {\em Proceedings of the
  IEEE conference on computer vision and pattern recognition}, pp.~589--597,
  2016.

\bibitem{sam2017switching}
D.~B. Sam, S.~Surya, and R.~V. Babu, ``Switching convolutional neural network
  for crowd counting,'' in {\em 2017 IEEE Conference on Computer Vision and
  Pattern Recognition (CVPR)}, pp.~4031--4039, IEEE, 2017.

\bibitem{zou2017occupancy}
J.~Zou, Q.~Zhao, W.~Yang, and F.~Wang, ``Occupancy detection in the office by
  analyzing surveillance videos and its application to building energy
  conservation,'' {\em Energy and Buildings}, vol.~152, pp.~385--398, 2017.

\bibitem{ren2015faster}
S.~Ren, K.~He, R.~Girshick, and J.~Sun, ``Faster r-cnn: Towards real-time
  object detection with region proposal networks,'' in {\em Advances in neural
  information processing systems}, pp.~91--99, 2015.

\bibitem{liu2016ssd}
W.~Liu, D.~Anguelov, D.~Erhan, C.~Szegedy, S.~Reed, C.-Y. Fu, and A.~C. Berg,
  ``Ssd: Single shot multibox detector,'' in {\em European conference on
  computer vision}, pp.~21--37, Springer, 2016.

\bibitem{krizhevsky2012imagenet}
A.~Krizhevsky, I.~Sutskever, and G.~E. Hinton, ``Imagenet classification with
  deep convolutional neural networks,'' in {\em Advances in neural information
  processing systems}, pp.~1097--1105, 2012.

\bibitem{pascal-voc-2012}
M.~Everingham, L.~Van~Gool, C.~K.~I. Williams, J.~Winn, and A.~Zisserman, ``The
  {PASCAL} {V}isual {O}bject {C}lasses {C}hallenge 2012 {(VOC2012)}
  {R}esults.''
  http://www.pascal-network.org/challenges/VOC/voc2012/workshop/index.html.

\bibitem{lin2014microsoft}
T.-Y. Lin, M.~Maire, S.~Belongie, J.~Hays, P.~Perona, D.~Ramanan,
  P.~Doll{\'a}r, and C.~L. Zitnick, ``Microsoft coco: Common objects in
  context,'' in {\em European conference on computer vision}, pp.~740--755,
  Springer, 2014.

\bibitem{seidel2018improved}
R.~Seidel, A.~Apitzsch, and G.~Hirtz, ``Improved person detection on
  omnidirectional images with non-maxima suppression,'' {\em arXiv preprint
  arXiv:1805.08503}, 2018.

\bibitem{tamura2019omnidirectional}
M.~Tamura, S.~Horiguchi, and T.~Murakami, ``Omnidirectional pedestrian
  detection by rotation invariant training,'' in {\em 2019 IEEE Winter
  Conference on Applications of Computer Vision (WACV)}, pp.~1989--1998, IEEE,
  2019.

\bibitem{cao2017realtime}
Z.~Cao, T.~Simon, S.-E. Wei, and Y.~Sheikh, ``Realtime multi-person 2d pose
  estimation using part affinity fields,'' in {\em CVPR}, 2017.

\bibitem{cao2018openpose}
Z.~Cao, G.~Hidalgo, T.~Simon, S.-E. Wei, and Y.~Sheikh, ``Open{P}ose: realtime
  multi-person 2{D} pose estimation using {P}art {A}ffinity {F}ields,'' in {\em
  arXiv preprint arXiv:1812.08008}, 2018.

\end{thebibliography}
\bibliographystyle{ieeetr}

\end{document}